\documentclass[conference]{IEEEtran}
\IEEEoverridecommandlockouts
\usepackage{cite}
\usepackage{amsmath,amssymb,amsfonts}
\usepackage{algorithmic}
\usepackage{graphicx}
\usepackage{textcomp}
\usepackage{xcolor}
\usepackage{subcaption}
\usepackage{enumitem}
\usepackage{algorithm}
\usepackage{algorithmic}
\usepackage[T1]{fontenc}

\def\BibTeX{{\rm B\kern-.05em{\sc i\kern-.025em b}\kern-.08em
    T\kern-.1667em\lower.7ex\hbox{E}\kern-.125emX}}

\renewcommand{\baselinestretch}{0.88}
\begin{document}

\title{Adaptive Meta-learning-based Adversarial Training for Robust Automatic Modulation Classification\\

\thanks{This material is based upon work supported by the National Science Foundation under Grant Numbers  CNS-2202972, CNS-2232048, and CNS-2318726.}
}

\author{
	\IEEEauthorblockN{
        Amirmohammad Bamdad\IEEEauthorrefmark{1}, 
	Ali Owfi\IEEEauthorrefmark{1}, 
        Fatemeh Afghah\IEEEauthorrefmark{1}}

    \IEEEauthorblockA{\IEEEauthorrefmark{1}Holcombe Department of Electrical and Computer Engineering, Clemson University, Clemson, SC, USA \\
        Emails: \{abamdad, aowfi,  
        fafghah\}@clemson.edu}

}

% \and
% \and

% }

\newcommand{\blue}[1]{\textcolor{blue}{#1}}
\newcommand{\red}[1]{{\textcolor{red}{#1}}}
\newcommand{\green}[1]{\textcolor{green}{#1}}
\newcommand{\orange}[1]{\textcolor{orange}{#1}}
\newcommand{\fa}[1]{\textcolor{magenta}{Fatemeh:#1}}
\newcommand{\ali}[1]{\textcolor{brown}{Ali:#1}}
\newcommand{\linke}[1]{\textcolor{brown}{#1}}
\newcommand{\yellow}[1]{\colorbox{yellow}{#1}}

\maketitle

\begin{abstract}
DL-based automatic modulation classification (AMC) models are highly susceptible to adversarial attacks, where even minimal input perturbations can cause severe misclassifications. While adversarially training an AMC model based on an adversarial attack significantly increases its robustness against that attack, the AMC model will still be defenseless against other adversarial attacks. The theoretically infinite possibilities for adversarial perturbations mean that an AMC model will inevitably encounter new unseen adversarial attacks if it is ever to be deployed to a real-world communication system. Moreover, the computational limitations and challenges of obtaining new data in real-time will not allow a full training process for the AMC model to adapt to the new attack when it is online. To this end, we propose a meta-learning-based adversarial training framework for AMC models that substantially enhances robustness against unseen adversarial attacks and enables fast adaptation to these attacks using just a few new training samples, if any are available. Our results demonstrate that this training framework provides superior robustness and accuracy with much less online training time than conventional adversarial training of AMC models, making it highly efficient for real-world deployment.
\end{abstract}


\section{Introduction}

% \ali{I personally think the interdiction is a bit lengthy now after the merge with related work. How about we delete the first paragraph and start the story from the second one?}
% One of the most researched applications of DL-based solutions in wireless communications is automatic modulation classification (AMC). AMC involves identifying the modulation scheme of received signals, a crucial task for spectrum management, interference recognition and mitigation, and cognitive radio systems. Theoretically, maximum likelihood provides the Bayes-optimal solution for modulation classification, but it assumes prior knowledge of channel conditions, which are highly dynamic in real-world communication systems. DL models were proposed as an alternative solution for AMC \cite{o2017introduction} which can implicitly learn channel variations and adapt to them given sufficient amount of data. Since then, various DL architectures such as CNNs\cite{west2017deep}, DNNs \blue{cite}, and LSTMs \blue{cite} \fa{define acronyms} have been proposed and evaluated for AMC, demonstrating the superior performance of DL-based AMC models over traditional methods. 

One of the most researched applications of deep learning (DL) in wireless communications is automatic modulation classification (AMC). Despite the impressive performance of DL-based AMC models\cite{o2017introduction} and the considerable research papers focusing on this field, their widespread adoption in practical systems remains limited. A key challenge lies in the vulnerability of DL models to adversarial attacks, a concern that extends beyond wireless communications and affects DL applications in general. Adversarial attacks involve subtle perturbations to input data that can cause significant degradation in model performance.

In the context of AMC, sources of interference using adversarial perturbations can severely undermine the accuracy of AMC models, and many researchers have pointed out this vulnerability \cite{sadeghi2018adversarial, lin2020threats, flowers2019evaluating}. These works have demonstrated that adversarial perturbations require significantly less power than AWGN to void AMC predictions. A couple of methods have been proposed as defensive mechanisms for AMC models against adversarial perturbations, such as randomized smoothing and data augmentation \cite{kim2021channel}, ensemble learning \cite{sahay2021deep}, and miss-classification correction \cite{sahay2021deep}. But arguably, adversarial training has been the most favored solution \cite{maroto2022safeamc,sahay2021robust}.

Although adversarial training—where the model is trained with adversarially perturbed examples—can significantly enhance robustness to observed attack types\cite{sahay2021robust,maroto2022safeamc}, the adversarially trained AMC model will remain vulnerable to new attack variations, e.g., attack perturbation not used in the adversarial training process. As theoretically, there can be infinite attack variations, an AMC model should expect to encounter unseen types of attacks when deployed to a real-time online system. This highlights the need for resilient DL-based AMC models that rely on fast adaptation to unseen adversarial attacks rather than DL models relying on adversarial training using a handful of pre-defined attack variations.

Moreover, a common yet unrealistic implicit assumption within the AMC literature is that the distribution of signals' data remains stationary \cite{flowers2019evaluating,maroto2022safeamc,kim2021channel}. These proposed AMC methods split a single dataset into a train-set and testing set and evaluated the model on the testing set that was sampled from the same distribution as the training set. This approach implicitly assumes that the wireless environment remains stationary over time. This is while the radio spectrum is highly dynamic and rapidly evolving in reality, demanding continuous adaptation and fine-tuning of the AMC model. Furthermore, most works also assume access to large amounts of annotated training samples, as well as ample training time and resources for an AMC model to train and adjust itself to wireless variations. In practice, however, AMC models are typically constrained by the limited availability of training samples and computational resources in real-time online scenarios. Limitations in the availability of training samples and training resources often result in a few-shot learning scenario, a factor largely neglected in the current AMC literature \cite{flowers2019evaluating,maroto2022safeamc,kim2021channel,o2017introduction,west2017deep}. 

\begin{figure}{}
        \centering
        \includegraphics[width=0.6\columnwidth]{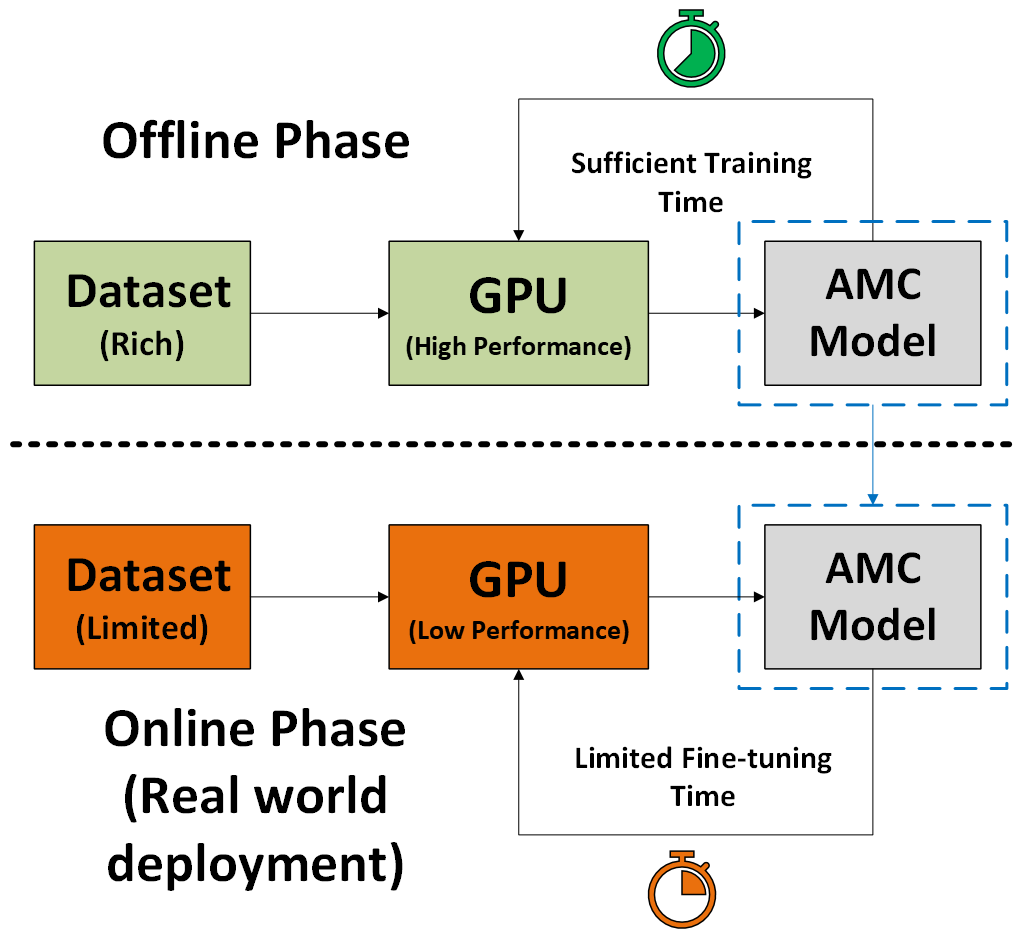} % Replace with your image
        \caption{Limitations of AMC models after being deployed to a practical real-world system (online-phase).}
        \label{fig:AMC-offline-online}
        \vspace{-15pt}
 \end{figure}

We believe a more realistic depiction of how AMC models can be practically implemented aligns with the system model shown in Fig \ref{fig:AMC-offline-online}. In the offline phase, sufficient datasets, powerful computational resources, and ample training time are available for pre-training and preparing the AMC model. Most previous AMC papers only consider this offline phase. However, we additionally consider an online phase where the pre-trained model is deployed within a practical system. During the online phase, the AMC model has to adapt and fine-tune itself in response to dynamic variations in real-time. Unlike the offline phase, the AMC model is challenged with limited new training samples, training resources, and training time in the online phase.

To address these limitations in prior work, we propose a meta-learning-based adversarial training framework for AMC models aimed at robustness against new adversarial attacks in the online phase. Unlike conventional adversarial training for AMC models, which focuses on resilience to specific pre-defined attack types, our approach equips the AMC model to learn a generalized adaptation strategy, which enables the model to rapidly adapt to new adversarial attacks using only a handful of adversarial samples. Even when there are no new training samples in the online phase, an AMC model can reach much better generalization in robustness against new adversarial attacks when our proposed meta-learning-based training framework is employed. Moreover, our proposed framework's low computational and minimal training sample requirements in the online phase satisfy the constraints of deploying an AMC model in real-time within a practical system. 

Meta-learning has been proposed in more recent years as an alternative to the conventionally trained DL models in various wireless applications such as interference identification \cite{owfi2023meta}, channel autoencoders \cite{park2020meta}, and classifying new modulation classes in AMC \cite{hao2024meta, hao2023automatic} as well. In one of the recent works \cite{zhang2022spectrum}, it is claimed that meta-learning is used to improve the robustness of AMC to black-box attacks. Based on the pseudo-code and the objective functions provided in the paper, their proposed method is multi-task learning and not meta-learning, as the loss function simply uses the average loss of the model on multiple tasks. \textit{To the best of our knowledge, we are the first to propose an adversarial training framework for AMC capable of fast adaptation to new adversarial attakcs in the online phase using meta-learning.} In summary, the main contributions of this paper are as follows:

\begin{itemize}
    
    \item Addressing the common unrealistic assumption of a single training/testing phase in AMC models by proposing a two-phased offline training/online deployment scenario for AMC models.
    \item Proposing a practical adversarial training framework for AMC models, more generalized in robustness against new adversarial attacks and capable of fast adaptation to new adversarial attacks in the online phase.
    \item Substantially increasing efficiency by lowering data and training requirements during online deployment.
      
\end{itemize}

% The rest of this paper is structured as follows: Section
% II discusses previous papers focusing on adversarial attacks in DL-based AMC models. Section III \blue{explain methodology}. In Section IV, experiment settings are explained and evaluations are provided.
% Finally, Section V concludes the paper.
% \input{2_related_work}
\section{Methodology} \label{method}

\subsection{Adversarial Attacks}
We first introduce the concept of adversarial attacks, as well the adversarial attack methods that we are using in our evaluations.
  Adversarial attacks seek to mislead models by adding small perturbations $\delta$ to inputs. Given a model \( f \), an input \( x \), and its true label \( y \), the goal is to maximize the model's error by altering x slightly while keeping $\delta$ small.

\begin{equation}
\begin{aligned}
&\arg \max_{\delta} \mathcal{L}(f(x + \delta), y) 
\\
&\text{s.t.} \quad \|\delta\|_p \leq \epsilon \quad \text{and} \quad x+\delta \in X
\label{eq:adv_attack}
\end{aligned}
\end{equation}
where $\|\delta\|_p$ denotes the $l_{p}$ norm of the perturbation. In the context of wireless communications, $l_{2}$ norm is the most natural and common norm to be used as it corresponds to the signal power. It should also be noted that $\delta$ is not unique. Solving for the optimal $\delta$ is not easy, and in most cases, sub-optimal perturbations can provide a favorable trade-off between performance and computation efficiency. Hence, various methods have been proposed that approximate the adversarial perturbation and are commonly used to apply adversarial attacks on DL models. We will briefly introduce the attacks that are used in this paper: 

\begin{itemize}[left=0pt, noitemsep]

    \item \textbf{Fast Gradient Sign Method (FGSM) Attack:} FGSM\cite{goodfellow2014explainingFSGM} is a simple and efficient adversarial attack that generates perturbations using the gradient of the loss function. 
\begin{equation}
\begin{aligned}
\delta = \epsilon \cdot \text{sign}(\nabla_x \mathcal{L}(f(x), y))
\label{eq:FSGM}
\end{aligned}
\end{equation}

The adversarial perturbation $\delta$ is created by taking a single step in the direction of the sign of the gradient of the loss with respect to the input $x$. The perturbation is scaled by a small factor $\epsilon$, ensuring that the resulting perturbation remains small enough to satisfy the constraints in  \eqref{eq:adv_attack} while maximizing the model's loss.

\item \textbf{Projected Gradient Descent (PGD) Attack:}
PGD is essentially an iterative version of FGSM that refines the perturbation over multiple steps\cite{mkadry2017towardsPGD,kurakin2016adversarialPGD}.  FGSM is repeatedly applied to update the perturbation while the perturbed input in each step is projected back into the allowed perturbation region (constrained by $l_p$-norm to satisfy the constraint in Equation \ref{eq:adv_attack}. 
\begin{equation}
x^{t+1} = \text{Proj}_{\epsilon} \left( x^t + \alpha \cdot \text{sign}(\nabla_x \mathcal{L}(f(x^t), y)) \right)
\end{equation}

This method is generally more effective than FGSM because it can escape local minima, leading to stronger adversarial examples that can fool models more reliably.

\item\textbf{Momentum Iterative Method (MIM) Attack:}
The Momentum Iterative Method (MIM) \cite{dong2018boostingMIM} builds upon the principles of iterative adversarial attack methods by incorporating momentum to improve the optimization process. The updating process at step t+1 for momentum $g^{t+1}$ and the perturbed input $x^{t+1}$ is defined as:
\begin{equation}
\begin{aligned}
g^{t+1} &= \mu \cdot g^t + \frac{\nabla_x \mathcal{L}(f(x^t), y)}{\|\nabla_x \mathcal{L}(f(x^t), y)\|_1} \\
x^{t+1} &= x^t + \alpha \cdot \text{sign}(g^{t+1})
\label{eq:MIM}
\end{aligned}
\end{equation}
where $\mu$ is the momentum coefficient and $\alpha$ is the step size. By utilizing momentum in its updates, MIM enhances the robustness of the generated adversarial perturbations, making them more transferable and effective in black-box attacks.

\item\textbf{Carlini and Wagner (C\&W) Attack:}
The Carlini and Wagner (C\&W) attack \cite{carlini2017towardsCW} is an adversarial attack that creates perturbed data by optimizing an objective function that combines two primary goals. First, it maximizes the model’s prediction error for the original class by pushing the data toward another class. Second, it ensures that the perturbed data remains as close as possible to the original by minimizing the norm of the perturbation. In this work, we use the $l_{2}$ norm, as discussed in the original C\&W paper, to measure and control the perturbation size when creating adversarial examples.

\begin{equation}
\begin{aligned}
& \min_{\delta} \|\delta\|_2^2 + c \cdot \mathcal{L}(f(x + \delta), y) \\
& \text{s.t.} \quad x + \delta \in [0, 1]^n
\end{aligned}
\end{equation}

\item\textbf{Principal Component Analysis (PCA) Attack:}
This method \cite{sadeghi2018adversarial} uses PCA to craft adversarial examples. It first calculates the principal components of a matrix consisting of the normalized gradients of the data. Since the first principal component captures the greatest variance in the data \cite{sadeghi2018adversarial}, perturbations are applied along this direction to disrupt the original data’s patterns in an effective manner.

\end{itemize}

\subsection{Problem Formulation and Design}

We are considering a DL-based AMC model which we denote as $f_{\theta}(.): \mathbb{R}^{2\times\lambda} \rightarrow \mathbb{R}^{C}$, where $\theta$ is the model's parameters, $2\times\lambda$ represents the dimension of input signals (in-phase and quadrature components for $\lambda$ timesteps), and $C$ is the number of possible modulation constellations. Given a received signal $y$, the AMC model predicts the modulation class based on
$ \arg \max_{i} \, \frac{\exp(f_{\theta}(y)_i)}{\sum_{k=1}^{C} \exp(f_{\theta}(y)_k)}$. 

\begin{figure}{}
        \centering
        \includegraphics[width=0.7\columnwidth]{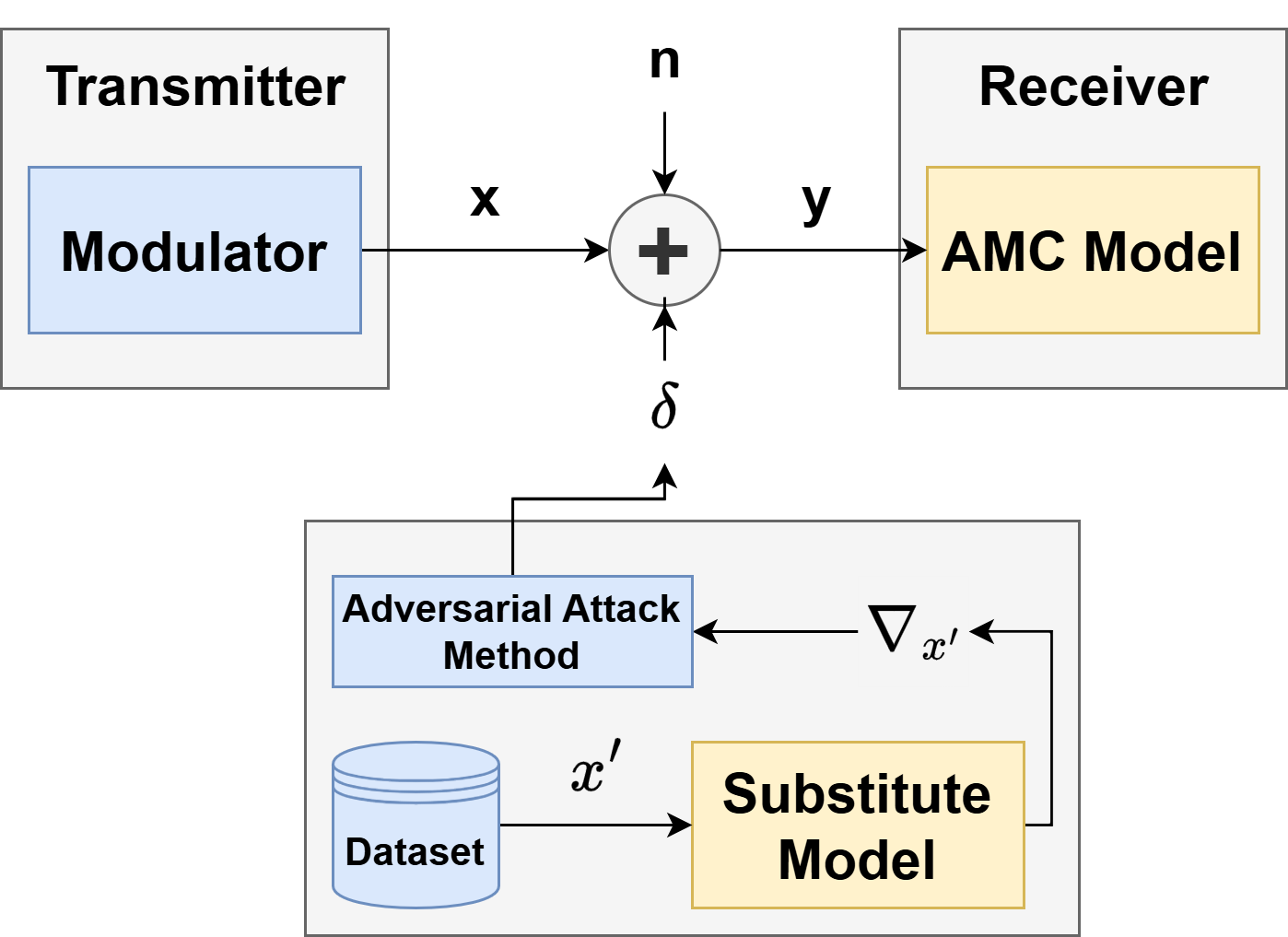} % Replace with your image
        \caption{Black-box adversarial attack on an AMC models.}
        \label{fig:adversarial_interference}
        \vspace{-15pt}
 \end{figure}

The AMC model is trained in the offline phase, where we do not have strict limitations on dataset, training time, and training resources. As shown in Fig \ref{fig:AMC-offline-online}, after the offline phase, the AMC model is deployed to a real-world system, which we call the online phase. During the online phase, the wireless environment evolves and differs from the offline phase, and hence, the AMC model has to adapt itself to the changes. In particular, as shown in Fig \ref{fig:adversarial_interference}, we are considering a wireless communication scenario with a source of interference utilizing an unknown adversarial attack to target the AMC model in the receiver specifically.

In this scenario, the received signal can be formally written as:
$y=H_{t}x+H_{a}\delta+n$.
The adversarial attack methods we introduced in the previous subsection require the gradient of the loss function with respect to the given input. As the source of interference is most likely not going to have direct access to the AMC model in the receiver, it will use a substitute AMC model instead to calculate the gradient and the adversarial perturbation; hence, the scenario becomes a black-box adversarial attack. Even though the generated perturbation is created such that it maximizes the error on the substitute AMC model, it is still devastating to the target AMC model as adversarial perturbations are transferable across various DL models to an extent\cite{szegedy2013intriguing}. 

 Note that while the AMC model could have experienced adversarial training in the offline phase, the adversarial attacks it faces in the online phase will almost definitely be new as there are infinite variations of adversarial perturbations. Thus, the AMC model remains susceptible to the adversarial attacks in the online phase. Adding to the challenges is that collecting adversarial training samples from new attacks in real-time may be infeasible, or at best, only a few samples can be obtained. Moreover, since the AMC model has to operate in real-time during the online phase, extensive training is also limited.

\subsection{Proposed Framework}
\begin{figure*}{}
        \centering
        \includegraphics[width=0.9\textwidth]{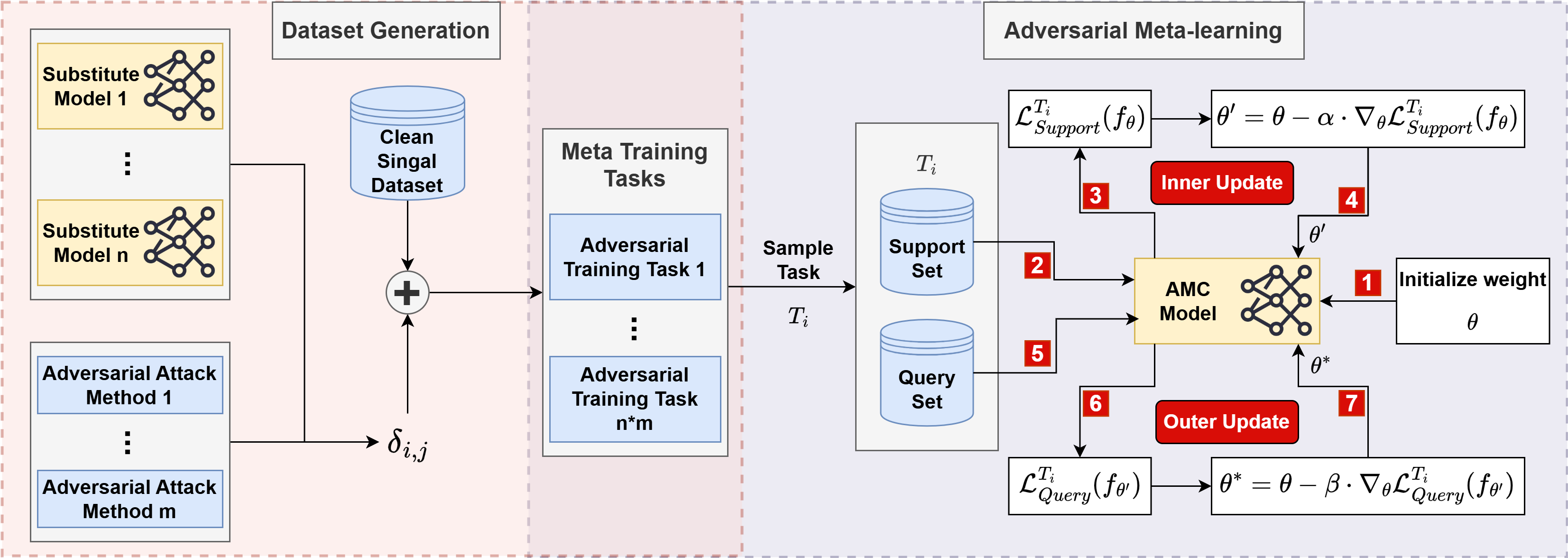} % Replace with your image
        \caption{Proposed meta-learning-based adversarial training framework for AMC models. The numbers in the adversarial meta-learning part denote the order of steps. The adversarial meta-learning section is depicted based on MAML, but theoretically, any model-agnostic meta-learning algorithm can be used as well. }
        \label{fig:adv_meta_amc}
        \vspace{-15pt}
 \end{figure*}

 \begin{algorithm}
\caption{Proposed adversarial training framework for AMC models}
\begin{algorithmic} 
\REQUIRE $\mathcal{A} = \{a_1, a_2, ..., a_n\}$ $\mathcal{S} = \{s_1, s_2, ..., s_m\}$, a set of adversarial attack methods and a set of substitute models 
\REQUIRE Clean signal dataset $\mathcal{D} = \{(x_k, y_k) \mid k = 1, \dots, N\}$ where $x_k$ is the received signal and $y_k$ is the modulation class

\STATE Randomly initialize weights $\theta$ for AMC model $f$ 
\FORALL{outer loop iterations}

\STATE Sample $a_i$ from $\mathcal{A}$ and $s_j$ from $\mathcal{S}$
\STATE Train $s_j$ using $\mathcal{D}$ 
\STATE Generate perturbation $\delta_{i,j}$ based on $a_i(s_j, \mathcal{D})$ 
\STATE $\mathcal{D}_{i,j} = \{(x_k+\delta_{i,j}, y_i) \mid k = 1, \dots, N\}$
\STATE Split $\mathcal{D}_{i,j}$ samples to $support$ and $query$ sets
\STATE Set $\theta' = \theta$

\FORALL{inner loop iterations}
\STATE Using $D_{i,j}$'s $support$ set compute loss $\mathcal{L}^{D_{i,j}}_{support}(f_{\theta'})$
\STATE Update $\theta^{'} \leftarrow \theta^{'} - \alpha\nabla_{\theta^{'}}\mathcal{L}^{D_{i,j}}_{support}(f_{\theta'})$
\ENDFOR

\STATE Using $D_{i,j}$'s $query$ set compute loss  $\mathcal{L}^{D_{i,j}}_{query}(f_{\theta'})$

\STATE Update $\theta \leftarrow \theta-\beta\nabla_{\theta}\mathcal{L}^{D_{i,j}}_{query}(f_{\theta'})$

\ENDFOR

\end{algorithmic}
\label{alg:META-adv}
\end{algorithm}

Our goal is to have a robust AMC model that can adapt to new adversarial attacks promptly when a limited number of new training samples are available and be generalized enough to remain robust to new attacks if it does not have access to any new training samples at all. Adversarially training the AMC model is the foundation of making the AMC model robust against adversarial attacks\cite{maroto2022safeamc}, but simply transferring an adversarially trained AMC model on attack $i$ to a scenario with a new attack $j$ will still result in a significant performance drop.

To tackle the extreme limitations during the online phase, we utilize an offline training framework for AMC models based on meta-learning adversarial training which enables transfer of knowledge from observed adversarial attack scenarios in the offline phase to new adversarial attacks in the online phase. Meta-learning, often referred to as learning to learn, is an approach in machine learning that aims to improve how quickly and effectively models can adapt to new tasks. This is achieved by utilizing knowledge learned from a set of different but similar tasks, known as meta-training tasks. By identifying common structures and patterns across these tasks, the model gains a foundational strategy for adaptation, allowing it to generalize to novel tasks, referred to as meta-test tasks, more efficiently. Meta-learning is especially useful for few-shot learning and limited-data scenarios, where the model’s ability to adapt quickly with minimal information is essential.

In meta-learning, each task is defined by a unique dataset. Every dataset captures the specific data distribution and environmental factors that distinguish one task from another.
Thus, by having distinct datasets, we effectively have distinct tasks. In the context of our AMC problem, each distinct task is
defined by the unique adversarial attack used by the source of interference affecting the AMC model. For each task, our goal is to adversarially train the AMC model to the unique adversarial attack. 

Based on this definition of task, the online phase adversarial attacks denote our meta-testing tasks. While we can not directly access these online phase adversarial attacks to fully adversarially train our model in the online phase, we can leverage different yet similar adversarial attacks to generate meta-training tasks to use during the offline phase. To create these meta-training tasks, we use a set of adversarial attack methods and a set of substitute models, as shown in Fig \ref{fig:adv_meta_amc}. Each combination of these substitute models and attack methods results in a set of unique adversarial perturbations. By injecting each set of the generated adversarial perturbations into the clean dataset, we generate a distinct adversarial training dataset and task.

After creating our meta-training tasks, we can utilize the adversarial meta-learning process. We will continue our explanations based on the Model Agnostic Meta-Learning (MAML) algorithm \cite{finn2017model}, but any other meta-learning algorithm that is also model agnostic, meaning that can be applied to any model architecture or learning approach without requiring specific modifications, can be used as well. We have additionally implemented our framework using two other meta-learning algorithms, namely Reptile\cite{nichol2018reptile} and FOMAML\cite{finn2017model}, which both use first-gradient approximations of the second-order gradient in MAML and are generally faster. 

At the start of the offline training, the AMC model $f$ is initialized with the weights $\theta$ (step 1 in Fig \ref{fig:adv_meta_amc}). The meta-learning process is divided into an inner loop and an outer loop stage. For each iteration in the outer loop, a meta-training task $T_i$ (or a batch of tasks) is sampled from the set of available meta-training tasks. Inside the outer loop, the inner loop is initiated by feeding $T_i$'s support set to the AMC model $f$ (step 2), calculating the loss $\mathcal{L}^{T_i}_{support}$ (step 3), and updating the weights to $\theta'$ (step 4) where:
$$\theta' = \theta - \alpha \cdot \nabla_{\theta} \mathcal{L}^{T_i}_{Support}(f_\theta) $$
After the inner loop, $T_i$'s query set is fed to the AMC model with the weights $\theta'$ (step 5), and the query set loss $\mathcal{L}^{T_i}_{query}$ is calculated (step 6). The initial weight $\theta$ is then updated (step 7) to $\theta^*$ where:
$$\theta^* = \theta - \beta \cdot \nabla_{\theta} \mathcal{L}^{T_i}_{Query}(f_{\theta'}) $$

After n outer loop iterations in the offline phase, $\theta^*$ becomes a strong initialization weight for the AMC model $f$, enabling small task-specific updates during inner-loop updates to quickly achieve high performance on new tasks. The pseudo-code for the offline training phase is also provided in Algorithm  \ref{alg:META-adv}. During the online phase, only the inner-loop is performed on the few-shot samples that may be available to the AMC model. 
\section{Results}

\subsection{Dataset Generation}

We use RML2016.10a\cite{o2016radioRML} dataset as the basis of the datasets we are generating, as it is a well-tested and publicly available dataset. The RML2016.10a dataset consists of 220000 samples in total. Each input sample is a 256-length vector, representing 128 in-phase and 128 quadrature signal components, and is linked to a distinct modulation scheme under a particular SNR. The dataset covers 20 SNR levels, ranging from -20 dB to 18 dB, increasing in 2 dB steps, and it includes 11 modulation types, BPSK, QPSK, 8PSK, QAM16, QAM64, CPFSK, GFSK, PAM4, WBFM, AM-SSB, and AM-DSB.

To generate the datasets contaminated with adversarial perturbations, we are considering eleven different neural network architectures as backbone models, namely, VTCNN, EfficientNet, VGG16, VGG19, ResNet18, MobileNet, MobileNetV2, ResNet50V2, ResNet50, DenseNet121, and Inception. We are also using five different adversarial attack methods, FGSM\cite{goodfellow2014explainingFSGM}, PGD\cite{kurakin2016adversarialPGD}, MIM\cite{dong2018boostingMIM}, C\&W\cite{carlini2017towardsCW}, and PCA, which we have introduced in Section \ref{method}. These backbone models act as the substitute models for the adversarial attack methods. We train each of these backbone architectures on the RML2016.10a dataset without any interferences. We then use the adversarial attack methods to generate perturbations on each of these trained AMC models. This results in 55 different perturbation sets which provides us with 55 unique datasets with interferences. From the perspective of meta-learning and transfer learning, each of these datasets denotes a separate \textit{task}. For all the meta-learning-based baselines, we are splitting the tasks into 50 meta-train tasks and 5 meta-test tasks.

\subsection{Baseline Models}

We are considering multiple AMC models in our experiments, each using a different offline training strategy, which we will introduce here:
\begin{itemize}[left=0pt, noitemsep]
    \item Scratch: This is a model that has not gone through any training in the offline phase, and is being trained from scratch during the online phase. This is not a viable option at all and is only being reported as an upper-bound for symbol error rate (SER).
    \item Transfer-Clean: The model has been fully trained on the clean dataset without any adversarial perturbations in the offline phase.
    \item Transfer-Adversarial: The model has been adversarially trained based on a specific attack. 
\end{itemize}

For fair comparison, we are using the same backbone neural network, VTCNN, for all of these models.

\subsection{Evaluations}

\subsubsection{Adaptation to Unknown Adversarial Attacks}

Even though adversarially training an AMC model using a specific adversarial perturbation will significantly increase its resistance against that adversarial perturbation, it will still remain weak against new and unknown types of perturbations. It is of key importance that the robust AMC model can adapt to new adversarial attacks as quickly as possible and with as few new training samples as possible.

\begin{figure}
    \centering
    % First subfigure
    \begin{subfigure}{\columnwidth}
        \centering
        \includegraphics[width=0.7\textwidth]{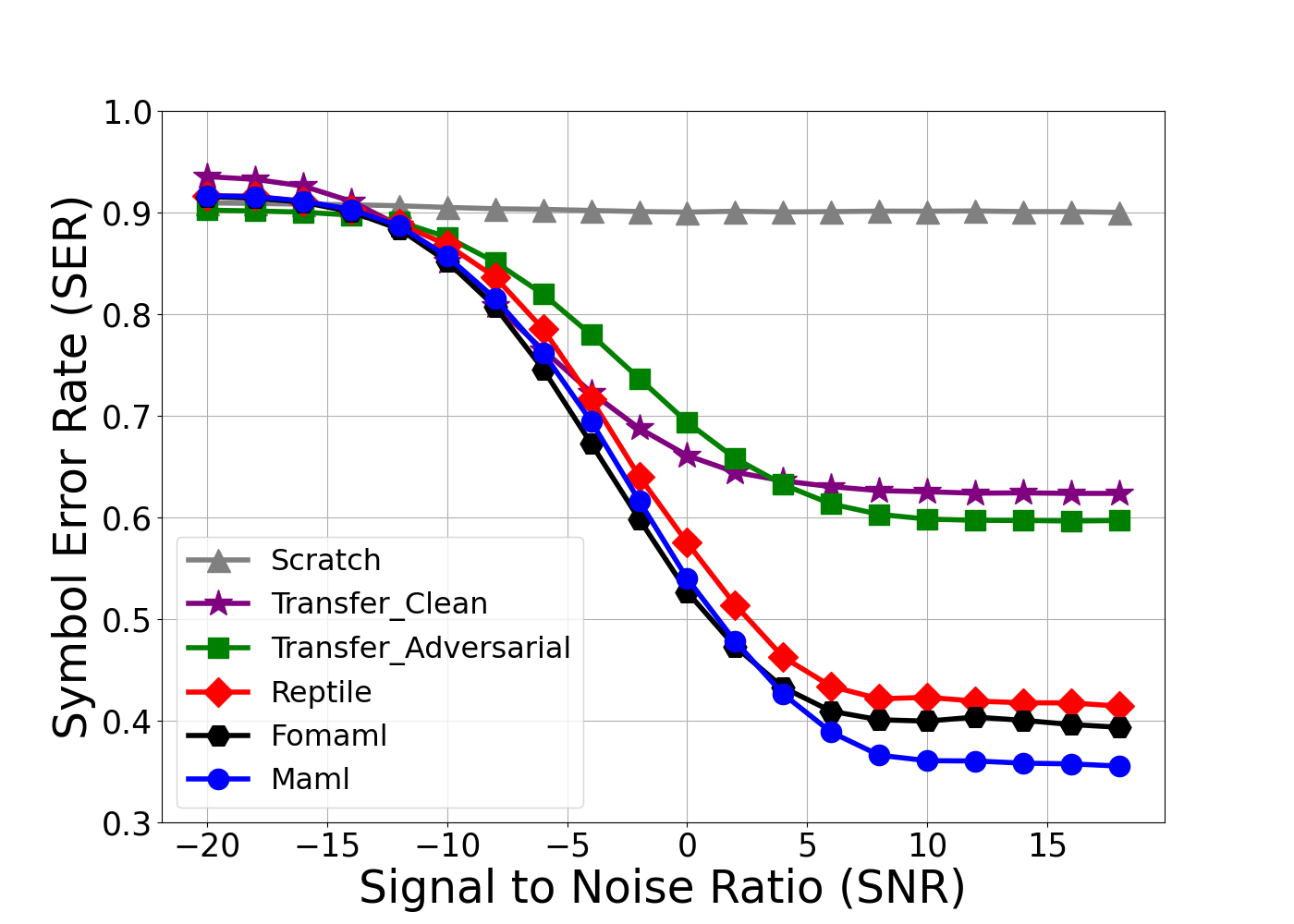} % Replace with your image
        \caption{SER in 2-shot scenario}
        \label{fig:first}
         \vspace{-15pt}

    \end{subfigure}
    
    \vspace{0.5cm} % Adjust the space between figures
    
    % Second subfigure
    % \begin{subfigure}{\columnwidth}
    %     \centering
    %     \includegraphics[width=0.99\textwidth]{plots/new5shot.png} % Replace with your image
    %     \caption{SER in 5-shot scenario}
    %     \label{fig:second}
    % \end{subfigure}
    
    \vspace{0.5cm} % Adjust the space between figures
    
    % Third subfigure
    \begin{subfigure}{\columnwidth}
        \centering
        \includegraphics[width=0.7\textwidth]{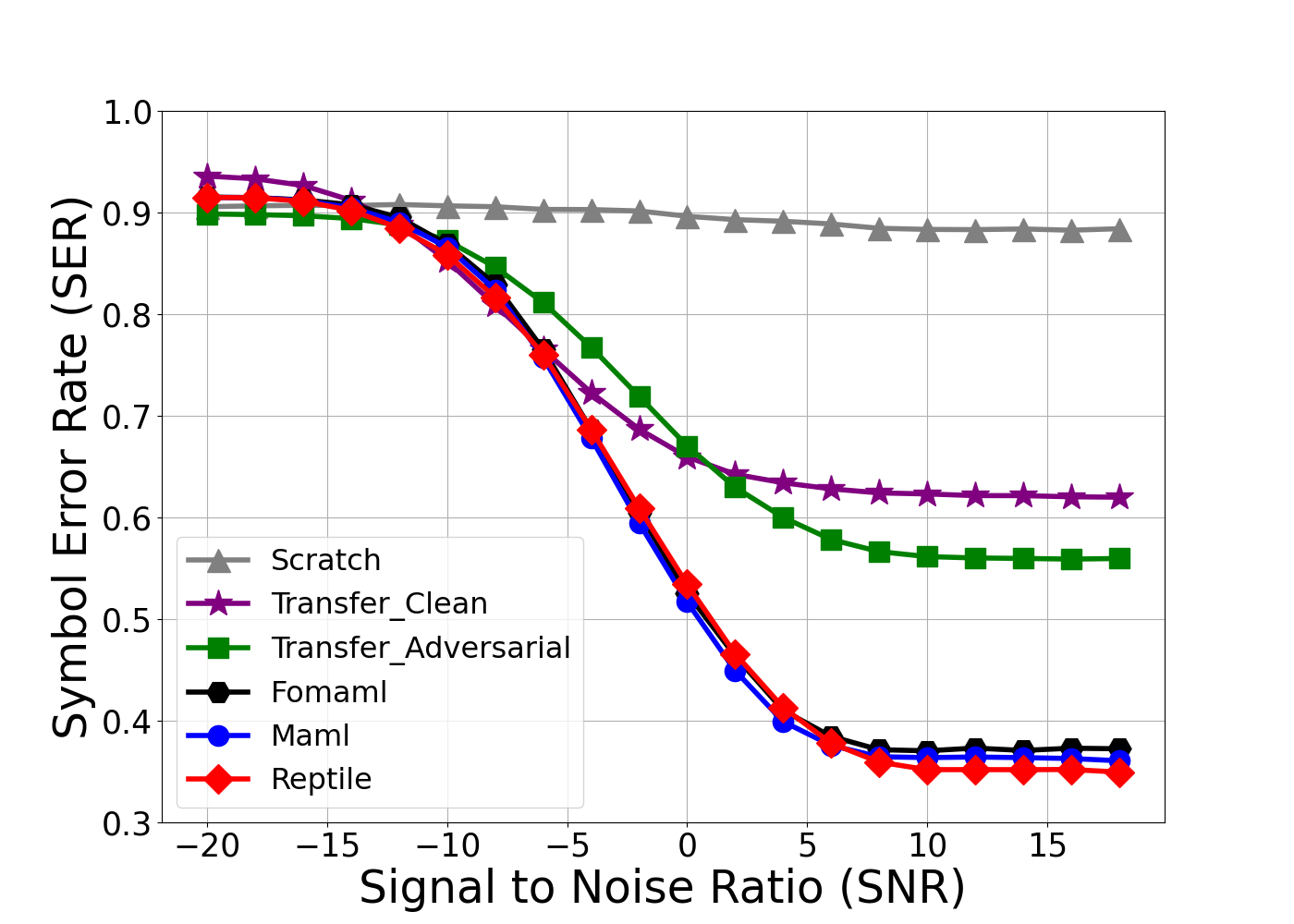} % Replace with your image
        \caption{SER in 10-shot scenario}
        \label{fig:third}

    \end{subfigure}
    
    \caption{Adaptation comparison of AMC baselines towards unseen adversarial attacks in few-shot scenarios. Only a few shots of adversarial training samples from the new tested adversarial attacks are provided to the AMC models.}
    \label{fig:SER_adaptation}
     \vspace{-15pt}

\end{figure}

To compare the adaptation capability of the AMC models, we have done few-shot learning experiments, provided in Fig \ref{fig:SER_adaptation}. In this experiment, each of the AMC models are tested against an unseen adversarial attack, and they have only access to a few shots of training samples from the new attack to adapt themselves to the new adversarial attack. The number of shots indicates the number of training samples available per modulation class.  

As it can be observed, the scratch model barely performs better than random in these challenging scenarios. While the Transfer-Adversarial performs better than Transfer-Clean, showing that adversarial training can contribute positively, it is still significantly outperformed by all our meta-learning-based baselines. This shows the advantage of meta-learning in enhancing the model's adaptability when it faces new attacks. Within the meta-learning-based AMC models, MAML provides the best performance in the more challenging 2-shot scenario. However, as the number of shots increases to 10, the Reptile AMC significantly improves and even slightly outperforms the more computationally expensive MAML AMC.

\subsubsection{Generalization to Unknown Adversarial Attacks}

\begin{figure}{}
        \centering
        \includegraphics[width=0.7\columnwidth]{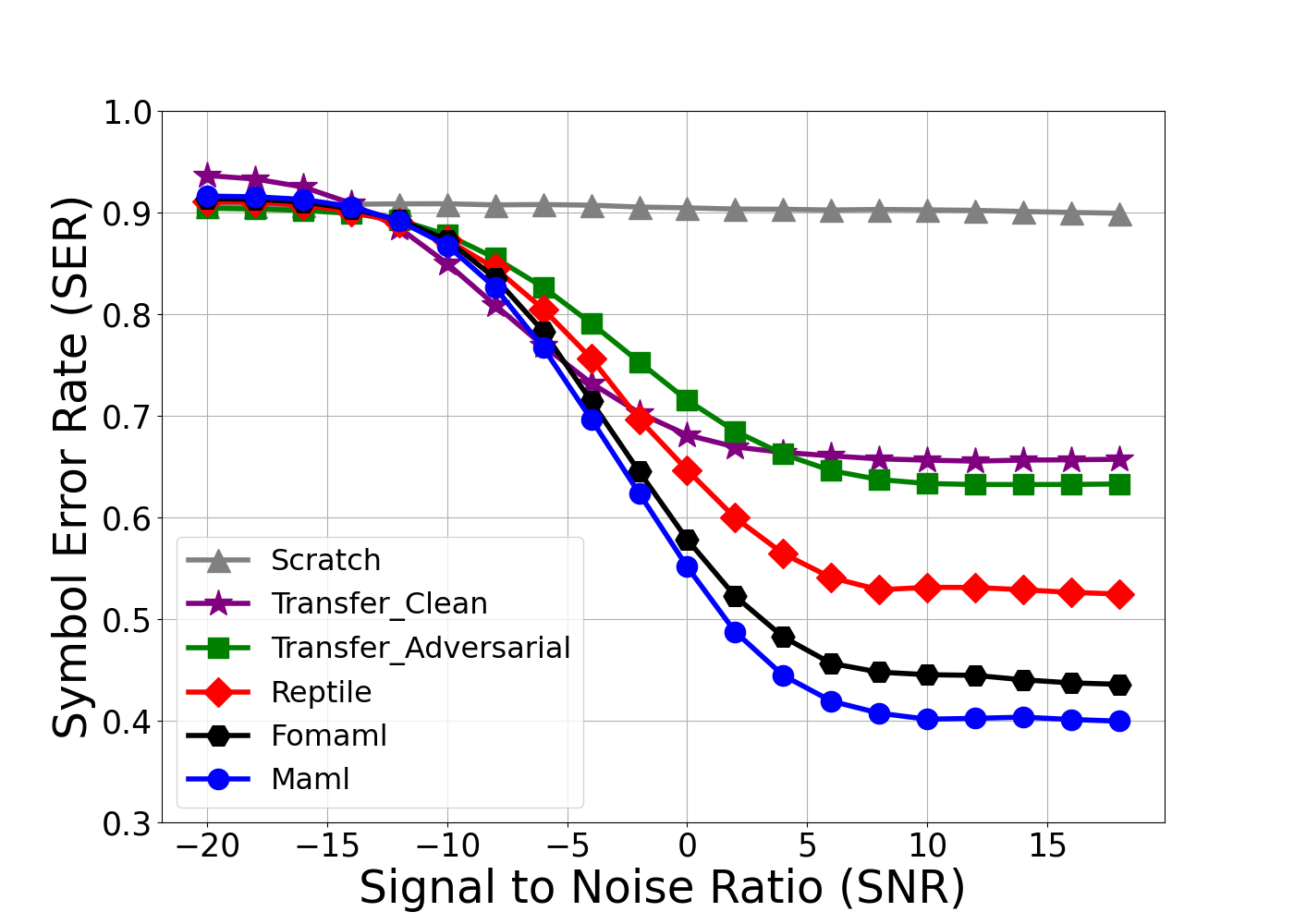} % Replace with your image
        \caption{Generalization comparison of AMC baselines against unseen adversarial attacks in 0-shot scenario. No new adversarial training samples from the new tested adversarial attacks are provided to the AMC models.}
        \label{fig:SER_generalization}
        \vspace{-15pt}
 \end{figure}

Another important aspect of AMC models' robustness against adversarial attacks is how resistant they are against unknown attacks without having access to any new training samples for fine-tuning. This capability is seen as generalization and translates into a 0-shot scenario in the context of DL models rather than a few-shot scenario. Figure \ref{fig:SER_generalization} compares the generalization of baseline AMC models against unknown adversarial attacks in the 0-shot scenario. The 0-shot scenario being an even more challenging version of the few-shot scenarios, Figure \ref{fig:SER_generalization} reinforces the findings in Figure \ref{fig:SER_adaptation}. All the AMC models following the proposed meta-learning-based training framework are substantially more robust against unseen adversarial attacks when no additional training samples are available. Among these AMC models, MAML has the best performance. It is worth noting that the performance of all meta-learning-based AMC models in 0-shot is still better than the performance of transfer learning baselines in 10-shot, as shown in Figure \ref{fig:SER_adaptation}. 

\subsubsection{Sample and Time Efficiency}

Two of the main limitations of DL-based AMC models when it comes to their integration in real-world practical systems, is firstly their need for large amounts of data, which may be infeasible to acquire in real-time, and secondly their training time and computation cost for fine-tuning the model online.  Other than increasing the robustness, utilizing the proposed meta-learning framework for AMC models will also significantly increase the efficiency both in terms of needed training samples, and online training time in real-time deployment.   

Table 1 provides the number of training shots needed from the new adversarial attack for each AMC model to roughly reach the same level of robustness. Meta-learning demonstrates a significant efficiency advantage, reaching the same robustness with 15, 30, and 160 times less samples compared to Transfer-Adversarial, Transfer-Clean, and Scratch AMC baselines respectively. This highlights the substantial improvement in shot efficiency achieved by meta-learning which mitigates the challenges of real-time deployment in practical systems with limited training samples.

% \begin{figure}{}
%         \centering
%         \includegraphics[width=0.98\columnwidth]{plots/2shot.jpg} % Replace with your image
%         \caption{Sample efficiency compassion among the AMC baselines during online fine-tuning. This figure demonstrates the numSER of shots (samples per modulation class) required for fine-tuning each AMC baseline to reach a specific SER under the an unseen adversarial black-box attack.}
%         \label{fig:efficiency}
%  \end{figure}
\begin{table}[ht]
\renewcommand{\arraystretch}{1.1} % Adjust row height
\centering
% \resizebox{\columnwidth/5}{!}{ 
\begin{tabular}{c|c|}
\cline{2-2}
                                      & \textbf{Samples per modulation class } \\ \hline
\multicolumn{1}{|c|}{Scratch} & 320                         \\ \hline
\multicolumn{1}{|c|}{Transfer-Clear}           & 60                       \\ \hline
\multicolumn{1}{|c|}{Transfer-Adversarial}        & 30                       \\ \hline
\multicolumn{1}{|c|}{Meta-learning-based}            & 2            \\ \hline
\end{tabular}
% }
\caption{Number of shots (samples per modulation class) needed for each of the AMC baselines from the new adversarial attack in the online phase to achieve the same accuracy in fine-tuning.}
\label{tab:sample_efficieny}
\end{table}

Offline and online training times for AMC models are provided in table \ref{tab:training_time}. The online training times are reported based on the number training samples each AMC model needs to reach the same SER for a new adversarial attack. While using the meta-learning framework increases the offline training time, it will significantly reduce the online training time. This is a favorable trade-off for us as real-time online performance is the more important factor. Moreover, utilizing light-weight meta-learning algorithms such as FOMAML or Reptile can also notably reduce the offline training time as well.

\begin{table}[ht]
\renewcommand{\arraystretch}{1.1} % Adjust row height
\centering
\resizebox{\columnwidth}{!}{ % Scale to fit within the column
\begin{tabular}{c|c|c|}
\cline{2-3}
                                      & \textbf{Offline Training} & \textbf{Online Training} \\ \hline
\multicolumn{1}{|c|}{Scratch} & -                         &                 11.60 sec        \\ \hline
\multicolumn{1}{|c|}{Transfer-Clear}           & 412 sec                     & 2.251 sec    \\ \hline
\multicolumn{1}{|c|}{Transfer-Adversarial}        & 417.5 sec                       &          1.432 sec                \\ \hline
\multicolumn{1}{|c|}{FOMAML}          & 600.636 sec                   &      0.419 sec                    \\ \hline
\multicolumn{1}{|c|}{Reptile}         & 608.994 sec                   &            0.406 sec               \\ \hline
\multicolumn{1}{|c|}{MAML}            & 899.30 sec            &          0.422 sec                 \\ \hline
\end{tabular}
}
\caption{Offline and online training time for the AMC models to reach the same accuracy in the online phase.}
\label{tab:training_time}
\end{table}

% \subsubsection{Offline training data efficiency of different meta-learning algorithms }

% \begin{figure}{}
%         \centering
%         \includegraphics[width=0.7\columnwidth]{plots/0shot.jpg} % Replace with your image
%         \caption{Task efficiency for }
%         \label{fig:efficiency}
%  \end{figure}
\section{Conclusion}

We have shown that even adversarially trained AMC models are still extremely susceptible to black-box attacks from unseen adversarial perturbations, especially in cases where there are little to no new training samples or training resources for the AMC model to fine-tune itself to the new adversarial attacks. We proposed a meta-learning framework for the offline training of AMC models, which we demonstrated can significantly increase robustness against unseen black-box adversarial attacks in terms of adaptation and generalization. Other than the significant increase in accuracy of the AMC model, using the proposed meta-learning framework will also substantially increase data and training resources efficiency in the online phase, allowing AMC models to operate in challenging online scenarios with data and computational limitations.

It should be noted that we proposed a new way of training AMC models to increase their robustness. Our suggested meta-learning framework and AMC training process are theoretically compatible with any backbone neural network architecture. For a fair comparison, we used VTCNN as the backbone for all the baselines in our experiments. Naturally, using a more complex backbone model should generally improve SER for all the baselines. We should note that our focus in this paper was not to find the best possible backbone model for AMC but rather to provide an AMC training framework that can be applied to any backbone model for improved results in robustness against unseen adversarial attacks.

\bibliographystyle{IEEEtran}
\bibliography{main}

\end{document}